\documentclass[11pt,english]{article}
\usepackage[utf8] {inputenc}    
\usepackage[arabic,english,french]{babel}
\usepackage{tia2013}
\usepackage{times}
\usepackage{latexsym}
\usepackage{amsmath}
\usepackage{graphicx}
\usepackage{amssymb}
\usepackage{float}
\usepackage{booktabs}
\usepackage{dcolumn}
\usepackage{makecell}
\usepackage{array,multirow,makecell}
\usepackage{url}

\usepackage{color}
\usepackage{xcolor}
\usepackage{colortbl}

\title{A Study of Association Measures and their Combination for Arabic MWT Extraction}

\author{Abdelkader El Mahdaouy \\
  LIM, Univ. USMBA \\
  Fès, Maroc \\
  Univ. Grenoble Alpes\\
  CNRS LIG/AMA \\
  Grenoble, France \\
  {\tt a.mahdaouy@hotmail.fr } \\\And
  Saïd EL Alaoui Ouatik  \\
  LIM,  Univ. USMBA \\
  Fès, Maroc \\
  {\tt s\_ouatik@yahoo.com } \\
  \And
  Eric Gaussier \\
  Univ. Grenoble Alpes\\
  CNRS LIG/AMA \\
  Grenoble, France \\
  {\tt eric.gaussier@imag.fr} \\}

\date{}
\begin{document}
\maketitle
\begin{abstract}
Automatic Multi-Word Term (MWT) extraction is a very important issue to many applications, such as information retrieval, question answering, and text categorization. Although many methods have been used for MWT extraction in English and other European languages, few studies have been applied to Arabic. In this paper, we propose a novel, hybrid method which combines linguistic and statistical approaches for Arabic Multi-Word Term extraction. The main contribution of our method is to consider contextual information and both termhood and unithood for association measures at the statistical filtering step. In addition, our technique takes into account the problem of MWT variation in the linguistic filtering step. The performance of the proposed statistical measure (NLC-value) is evaluated using an Arabic environment corpus by comparing it with some existing competitors. Experimental results show that our NLC-value measure outperforms the other ones in term of precision for both bi-grams and tri-grams.
\end{abstract}
\section{Introduction}
\paragraph{} Automatic Multi-Word Term extraction is an important task in many Natural Language Processing (NLP) applications \cite{Boula,Wen07}. The aim of the MWT acquisition process is to extract specific domain terms from special language corpora \cite{Kor08}. The extraction of MWTs is crucial for terminology acquisition, since they are less ambiguous and less polysemous than single word terms, and since their internal structure encodes useful semantic relations \cite{Wen08}.
\paragraph{}There are three main approaches to MWT extraction. The first one makes use of linguistic filters. The second one relies on statistical measures based on termhood and/or unithood. Termhood denotes {\em“the degree to which a linguistic unit is related to a specific domain concept”}, and unithood denotes {\em “the degree of strength or stability of syntagmatic combinations or collocations”} \cite{Kag96}. Lastly, the third approach is hybrid and combines the linguistic and the statistical approaches. Hybrid methods extract MWTs using linguistic filters and then rank the list of candidate MWTs according to statistical measures.
\paragraph{}In this paper, we propose a novel, hybrid method for Arabic MWT extraction. Like other hybrid methods, it includes two main filters. In the first one, we use a part-of-speech (POS) tagger to extract candidate MWTs based on syntactic patterns. In the second one, we propose a novel statistical measure, the NLC-value, that unifies the contextual information and both termhood and unithood measures. We compare this measure to alternative ones in the task of MWT extraction:  NTC-value \cite{Vu08}, LLR+C-value \cite{Khatib10}, C/NC-value and LLR.      
\paragraph{}The remainder of this paper is organized as follows. In the next section, Section 2, we present the related work. Section 3 describes the proposed method to extract MWTs. In Section 4, we present how MWT variation is handled in the proposed method. Section 5 describes the experimental validation and Section 6 concludes this work and presents some perspectives.
\section{Related Work}
\paragraph{}Several studies have been conducted on MWT extraction for many languages. These studies have either used a linguistic approach, a statistical approach, or a combination of them (hybrid approach). Most recent MWT extraction methods rely on a hybrid approach to efficiently extract MWTs, due to its higher accuracy compared to the two other approaches \cite{Tad03}.
 The linguistic approach uses technical analysis on the current knowledge of the language and its structure. There are two subcategories: approaches based on morpho-syntactic patterns \cite{daille94} and those based on MWT boundary detection \cite{Bour94}.
\paragraph{}The main purpose of applying statistical methods for MWT extraction is to rank candidate terms based on a particular measure that gives higher scores to "good" candidate terms. Candidate terms above a particular threshold are selected for further processing. The reliance on frequency is based on the simple assumption that a frequent expression indicates an important representation of the domain in question. Therefore, frequent expressions are assumed to represent important concepts. Given a candidate multi-word term, frequency only counts how often the candidate occurs in the text, but doesn’t give any information on the strength of the relationship between words composing the candidate multi-word term. Statistical approaches aim at extracting candidate terms from text corpora by means of association measures \cite{chu91} that concentrate on termhood and/or unithood to assign a score to candidate MWTs. These measures are based on frequency and co-occurrence information such as the T-score \cite{chu91}, the loglikelihood ratio (LLR) \cite{Dunning94}, the C/NC-Value \cite{Frantzi98}, etc.
\paragraph{}While linguistic approaches focus on syntactic structures, statistical methods focus on the recurrent characteristics of MWTs.  Both have their advantages and limitations. As mentioned by Boulaknadel et al. \shortcite{Boulb}, statistical approaches {\em “are unable to deal with low-frequency MWTs“} while pure linguistic approaches are {\em “language  dependent  and  not  flexible  enough  to  cope with  complex  structures  of  MWTs” }. Hybrid methods try to combine linguistic and statistical techniques to extract MWTs in order to avoid the weaknesses of the two approaches. 
\paragraph{} Boulaknadel et al. \shortcite{Boulb} have relied on a hybrid method to extract Arabic MWTs. As a first step, candidate terms that fit syntactic patterns are extracted from the output of the part-of-speech (POS) tagging tool proposed by Diab \shortcite{diab04}. In the second step, the list of candidate terms is ranked according to one of the following association measures: log-likelihood ratio (LLR), Mutual Information (MI), FLR, and T-score. These measures have been evaluated on an Arabic corpus and he results obtained show that LLR outperforms the other association measures. 
\paragraph{} Bounhass et al.\shortcite{Bounhas09} have followed the same approach (using again Diab's \shortcite{diab04} POS tagger and LLR) while focusing on compound nouns and thus using a more restricted set of syntactic patterns. For the bigrams, the obtained results outperform those obtained by Boulaknadel et al. \shortcite{Boulb}. 
\paragraph{} A similar study has been conducted by Al Khatib et al. \shortcite{Khatib10}, based on the POS tagger proposed by \cite{Taani09} and an association measure that combines both termhood and unithood through a combination of the C-value and the LLR. Experimental results show promising results for the combined measure.
\paragraph{}Most hybrid methods presented previously have been evaluated on 100 (best) candidate MWTs and deal with bi-grams (i.e. candidate MWTs of length 2). Moreover, they rely on LRR or a combination of LRR and C-value \cite{Khatib10} and ignore contextual information in the ranking step. To overcome this limitation, we introduce a new association measure that integrates contextual information and both termhood and unithood. Our overall approach is also hybrid and relies on the same linguistic filters as the ones used in the previous studies, based on syntactic patterns applied on the output of the POS tagger developed by \cite{diab09}.
\section{Proposed Method}
\paragraph{}Our method for extracting MWT candidates comprises two major steps: the linguistic and the statistical filters.
\subsection{Linguistic Filter}
\begin{figure}[!h]
\includegraphics[width=7.5cm,height=8cm]{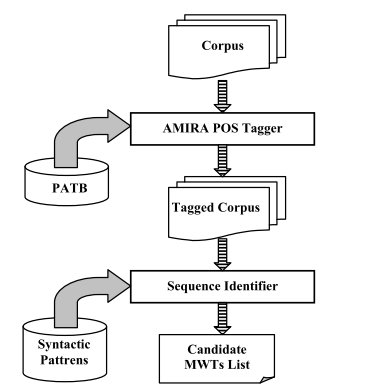} 
\caption{The global schema of the linguistic filter}
\label{lingFilter}
\end{figure}

\paragraph{}The proposed linguistic filter extracts candidate MWTs based on two core components; the POS tagger and the sequence identifier. In the literature, several methods for Arabic POS tagging systems have ben developed. We have used the one proposed by \cite{diab09} as it performs at over 96\% accuracy and allows a number of variable user settings. The underlying system uses Support Vector Machine (SVM). Figure (\ref{lingFilter}) illustrates the global schema of our linguistic filter.
As a first step, our method tags the corpus using the AMIRA toolkit (Diab POS Tagger) which is trained from the Penn Arabic TreeBank  (PATB) to assign tags for each word in the corpus. Then, the sequence identifier tokenizes tagged files of the corpus and uses syntactic patterns in order to identify candidate terms that fit the rules of the grammar. We have extended the list of syntactic patterns used by Boulaknadel et al. \shortcite{Boulb} as follows:
 \begin{itemize}
 \item $(Noun +(Noun|ADJ)+|(Noun|ADJ)+\\ |(Noun|ADJ))$
 \item $ Noun \, Prep \, Noun$
 \end{itemize}
   The second major step of the linguistic filter is handling the problem of MWTs variation   to improve the effectiveness  of extracted MWT candidates. Several categories of term variation are taken into account by this filter: graphical, inflectional, morpho-syntactic and syntactic variants, and are discussed in Section 4.
\subsection{Statistical Filter}
\paragraph{}In this step, we apply a number of statistical measures to rank the list of candidate MWTs extracted by the linguistic filter. The main objective of our statistical filter is to consider both termhood and unithood measures.
\subsubsection{The $C$-value}
\paragraph{}The $C$-value measures the termhood of a candidate string on the basis of several characteristics: number of occurrences, term nesting, and term length. It is defined as:
\begin{equation}
\label{cvalue}
\begin{split}
C\textrm {-Value}(a)=
\begin{cases}
\displaystyle
\log_{2}(|a|)\cdot f(a) \,\, \textrm { if  $a$ is not nested,}\\
\log_{2}(|a|)\cdot (f(a)-g(a))\,\,\textrm{otherwise}
\end{cases}
\end{split}
\end{equation}
where $|a|$ denotes the length in words of candidate term $a$, $f(a)$ is the number of occurrences of $a$ and:
\[
g(a)=\frac{1}{|T_{a}|}\sum_{b\in T_{a}}f(b)
\]
where $T(a)$ denotes the set of longer candidate terms into which $a$ appears ($|T(a)|$ is the cardinality of this set).

As one can note, if the candidate term is not nested, its score is solely based on its number of occurrences and length. If it is nested, then its number of occurrences is corrected by the number of occurrences of the terms into which it appears.


\subsubsection{The $NC$-value}
\paragraph{} The $NC$-value combines the contextual information of a term together with the $C$-Value. The contextual information is calculated based on the $N$value which provides a measure of the terminological status of the context of a given candidate term. It is defined as:
\begin{equation}
\label{nvalue} 
\displaystyle N\textrm{value (a)}=\sum_{b \in C_{a}}f_{a}(b)\cdot \frac{|T(b)|}{n}
\end{equation}
where $C_a$ denotes the set of distinct context words of $a$, $f_a(b)$ corresponds to the number of times $b$ occurs in the context of $a$ and $n$ is the total number of terms considered. This measure is then simply combined with the $C$-value to provide the overall $NC$-value measure:
\begin{equation} 
\label{ncvalue}
NC-\textrm{value(a)}=0.8\cdot C -\textrm{value(a)}+0.2\cdot N\textrm{value(a)}
\end{equation}

\subsubsection{The $NTC$-value}

\paragraph{}The aim of the $NTC$-value \cite{Vu08} is to incorporate a unithood feature, through the T-score, to the $C/NC$-value to improve its performance. The T-score measures the adhesion or differences between two words in a corpus of $N$ words as follows:
\begin{equation}
\label{tscore}
\displaystyle
T{s}(w_{i},w_{j})=\frac{p(w_{i},w_{j})-p(w_{i})\cdot p(w_{j})}{\sqrt{\frac{p(w_{i},w_{j})}{N}}}
\end{equation}
where $p(w_{i},w_{j})$ corresponds to the probability of observing the bi-gram $w_{i},w_{j}$ in the corpus; $p(w_{i})$ is the probability of word $w_i$ in the corpus and corresponds to the marginal probability of $p(w_{i},w)$. The T-score is integrated in the $C/NC$ measures through a re-weighting of the number of occurrences that privileges terms with a positive T-score:
\begin{equation}
\label{Freq}
F(a)=\begin{cases}
\displaystyle
f(a) \,\,\,\,\,\, \textrm {   if  $\min(T{s}(a))\leq 0$   } \\
f(a)\ln(2+\min(T{s}(a))\textrm{otherwise}
\end{cases}
\end{equation}
where $\min(T{s}(a))$ corresponds to the minimum T-score obtained from all the word pairs in $a$. Substituting $F(a)$ to $f(a)$ in Equation~\ref{cvalue} yields the $TC$-value, which is then combined with the $N$value as before, leading to the $NTC$-value:
%
\begin{equation}
\label{ntcvalue}
NTC\textrm{-value(a)}=0.8\cdot TC\textrm{value(a)}+0.2\cdot N\textrm{value(a)}
\end{equation}
The resulting metric (\ref{ntcvalue}) thus takes into account both contextual information and termhood and unithood measures.\\

\subsubsection{The $NLC$-value}
 
\paragraph{} We follow here the same development as before but rely this time on the more accurate unithood feature LLR \cite{Dunning94}, instead of the T-score, for the combination with the C/NC-value \cite{Frantzi98}. LLR is a "goodness of fit" statistics that determines if the words in an observed $n$-gram come from a sample that is independently distributed (meaning they co-occur by chance) or not. The underlying measure is calculated for bi-grams by the following formula:
\begin{align*}
&LLR(w_{j},w_{j}) =a\log(a)+b\log(b)+c\log(c) \\
&+d\log(d)-(a+b)\log(a+b)   \\
&-(a+c)\log(a+c)-(b+d)\log(b+d) \\
&-(c+d)\log(c+d)+N\log(N)
\end{align*}
with:\\
$a$: number of terms in which $w_i$ and $w_j$ co-occur;\\
$b$: number of terms in which only $w_i$ occurs;\\
$c$: number of terms in which only $w_j$ occurs;\\
$d$: number of terms in which neither $w_i$ nor $w_j$ appear;\\
$N$: total number of extracted terms.\\

For terms that consist of more than two terms, we calculate the LLR for each big-ram and then consider the minimum value obtained. The number of occurrences of a term is now re-weighted by this minimum value: $FL(a)=f(a)\cdot\ln(2+\min(LLR(a)))$ which is used instead of $f(a)$ in the $C$-value, leading to the $LC$-value:
\begin{equation}
\label{lcvalue}
L\textrm{C-value(a)}=\begin{cases}
\displaystyle
\log_{2}(|a|) \cdot FL(a) \,\, \textrm { if  $a$ is not nested,} \\
\log_{2}(|a|) \cdot (FL(a)-GL(a) ) \,\,\textrm{else}
\end{cases}
\end{equation}
with $GL(a)=\frac{1}{|T_{a}|}\sum_{b \in T_{a}}FL(b)$\\

This measure is then combined with the $N$value as before, leading to the $NLC$-value that integrates contextual information and both termhood and unithood:
\begin{equation}
\label{nlcvalue}
NLC\textrm{-value(a)}=0.8\cdot LC\textrm{-value(a)}+0.2\cdot N\textrm{value(a)}
\end{equation}

\section{Term variation}
\paragraph{} As mentioned in the previous section, we have handled the problem of term variation at the linguistic step. Our method takes into account four types of variations: graphical variants, inflectional variants, morpho-syntactic variants and syntactic variants.  Graphical variants concern orthographic errors occurred in writing a particular letters (\AR{”أ“}, \AR{”ي“} and \AR{”ة“}) which are very common in Arabic.  Furthermore, some letters go through a slight modification in writing, that doesn’t necessarily change the meaning of the word. For example, the letter \AR{”ي“} is replaced by another letter \AR{ ”ى“} at the end of a MWT, as for \AR{”التلوث الكيميائي“} which leads to \AR{“التلوث الكيميائى”} meaning “chemical pollution”. Inflectional variants are due to the use of different forms for the words constituting a MWT; these different forms are related to gender and number of adjectives, as in \AR{”تلوث المحيط“} (ocean pollution) and \AR{”تلوث المحيطات“} (pollution of the oceans) and to the presence/absence of a definite article, as in \AR{”تلوث مياه“} (water pollution) and \AR{”تلوث المياه“} (the water pollution). Morpho-syntactic variants affect the internal structure of term as the words it contains are related through derivational morphology. Two patterns control this type of variation in Arabic MWTs:
\begin{itemize}
\item $Noun1 \, Noun2 \Leftrightarrow Noun1 \, Adj$: \AR{”تلوث الهواء“} and \AR{”التلوث الهوائي“} (“air pollution”).
\item $Noun1 \, Adj \Leftrightarrow Noun1 \, Prep \, Noun$: \AR{”برميل نفطي“} and \AR{”برميل من النفط“} (“barrel of oil”).
\end{itemize}
We treat these three types of variations by using normalization method and the light stemming algorithm described in~\cite{Lark07} on each word of each MWT candidate. 

Syntactic variants modify the internal structure of the MWT candidate by adding one or more words (as adjectives) but do not affect the grammatical categories of the content words of the original MWT candidate. Such variants can be identified, for a given MWT candidate, by searching for all the stemmed MWT candidates that contain it. All the elements that constitute an addition to the original MWT candidate are then considered as context terms.

\section{Experiments and Results}

\subsection{The Corpus }

\paragraph{} Since there is no standard domain-specific Arabic corpus, we have built, in order to evaluate our approach, a new corpus specialized on the environmental domain with similar properties as the ones described \cite{Boulb,Bounhas09,Khatib10}. 
\paragraph{} The corpus built contain 1666 files comprising 53569 different tokens (without stop words) extracted from the Web site “Al-Khat Alakhdar”\footnote{http://www.greenline.com.kw}. It covers various environmental topics such as pollution, noise effects, water purification, soil degradation, forest preservation, climate change and natural disasters.

\subsection{Evaluation and Results }
\paragraph{} The evaluation of automatic MWTs extraction is a complex process and is usually performed by comparing each MWT candidate extracted to a domain-specific reference list. When there is no reference list available in the language retained, one can first translate the MWT candidates (using a machine translation system or a bilingual dictionary) and use a reference list available in another language. For the evaluation purpose, we have constituted automatically a reference list of all Arabic MWTs available in the latest version of AGROVOC\footnote{www.fao.org/agrovoc/} thesaurus and then use the stemming algorithm to remove prefixes and suffixes for each MWT in the reference list and the extracted MWT list. The next step consists of using an algorithm that considers a MWT candidate as correct if it is included in this list, noting that the MWT candidate and the term in the reference list should have the same number of stemmed words. Otherwise, we translate it and consider it as relevant whether its translation is contained in the European terminological database IATE\footnote{http://iate.europa.eu/iatediff}. Finally, the precision is calculated using the number of attested MWTs and the number of considred terms. 

\paragraph{}We computed the association scores (LLR, C-value, NC-value, NTC-value, LLR+C-value, NLC-value) for the MWT candidates and retain from each produced ranking for each statistical measure the $k$-best candidates, with $k$ ranging  from $100-300$ at intervals of $100$. 
 The experimental results illustrated in table \ref{tab2} show that our method (NLC-value) outperforms the previous methods in term of the quality of the extracted MWTs.
 
\begin{figure}[!h]
\includegraphics[width=7.5cm,height=5.3cm]{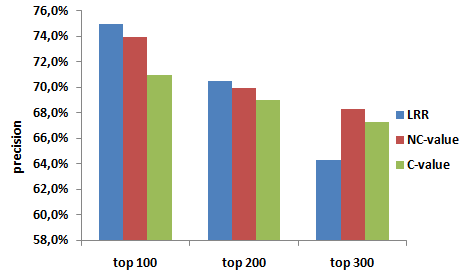} 
\caption{Precision obtained for the LLR and the $C/NC$-value}
\label{fig1}
\end{figure}

\begin{table}[!h]
\begin{center}
\renewcommand {\arraystretch }{1}
\begin{tabular}{@{}lllD{,}{,}{1}D{,}{,}{1}@{}}
\Xhline{1.0 pt}
 & \multicolumn{4}{l}{{\bf Top MWT considred}}  \\  
\cline{2-5}
\centering {\bf Stat. measures} & \centering {\bf 100} & \centering {\bf 200}& \centering {\bf 300} & \tabularnewline
\hline
{\bf LLR}   & 75,0\%  & 70,5\% & 64,3\%\\
{\bf C-value} & 71,0\%	& 69,0\% & 67,3\%\\
{\bf NC-value} &74,0\% & 70,0\%	& 68,3\%\\
{\bf NTC-value}& 80,0\% &	71,5\% &	69,7\%\\
{\bf LLR+C-value}   &73,0\% & 72,0\% & 68,3\%\\
{\bf NLC-Value}   &82,0\% & 75,5\% & 73,0\%\\
\Xhline{1.0 pt}
\end{tabular}
\end{center}
\caption{Results obtained for different statistical measures}
\label{tab2} 
\end{table}
\paragraph{} Furthermore, the combination of the context information and the $C$-value improves the performance of the process of MWT extraction because the $NC$-value outperforms the $C$-value for each considered MWT list. The unithood feature LLR outperforms the $C/NC$-value as expected from previous studies. Figure \ref{fig1} illustrates the precision obtained for the $C/NC$-value and the LLR.

\begin{figure}[!h]
\includegraphics[width=7.5cm,height=5.3cm]{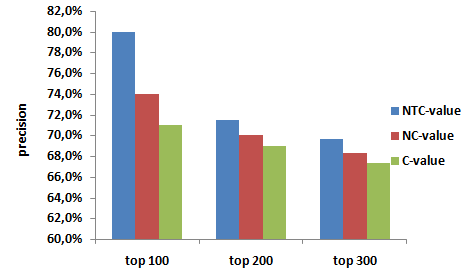} 
\caption{Precision obtained for the $C/NC$-value and the $NTC$-value}
\label{fig2}
\end{figure}

\begin{figure}[!h]
\includegraphics[width=7.5cm,height=5.3cm]{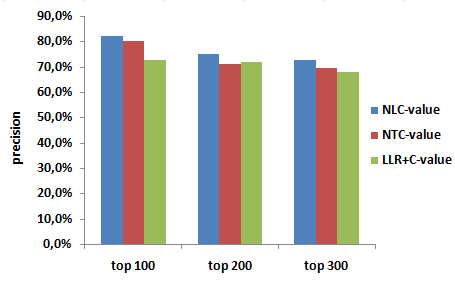} 
\caption{Precision obtained for different statistical measures that combine termhood and unithood}
\label{fig3}
\end{figure}

\paragraph{} The integration of contextual information and the T-score unithood measure to the $C$-value improves the performance of MWT acquisition, since the $NTC$-value has better precision than the $C/NC$-value, as illustrated in Figure \ref{fig2}.
\paragraph{} Lastly, the combination of termhood and unithood measures ($NTC$-value, $LLR+C$-value, $NLC$-value) is essential for MWT extraction, since all the measures based on this combination perform better than measures using only termhood or unithood ($C$-value, $NC$-value, LLR). We note that the statistical measure we have propose, $NLC$-value, outperforms all other measures. This measure is based on the accurate unithood feature LLR, combined with the $NC$-value. The $NLC$-value method takes advantages from previous works proposed in \cite{Vu08} and \cite{Khatib10} taken into account contextual information and both termhood and unithood association measures. Figure \ref{fig3} presents a comparaison of the precision for different statistical measures that combine termhood and unithood.

The number of different terms evaluated are 1095 amongst other 1800 terms, moreover the statical measures share 141 terms. The tables \ref{tab3} and \ref{tab4} represent the number of terms found in agrovoc and IATE respectively.
\begin{table}[!h]
\begin{center}
\renewcommand {\arraystretch }{1}
\begin{tabular}{@{}lllD{,}{,}{1}D{,}{,}{1}@{}}
\Xhline{1.0 pt}
 & \multicolumn{4}{l}{{\bf Top MWT considred}}  \\  
\cline{2-5}
\centering {\bf Stat. measures} & \centering {\bf 100} & \centering {\bf 200}& \centering {\bf 300} & \tabularnewline
\hline
{\bf LLR} &  35	& 60 & 80\\
{\bf C-value} & 27 & 59 & 82\\
{\bf NC-value} & 32 & 62 & 82\\
{\bf NTC-value}& 35 & 60 & 83\\
{\bf LLR+C-value}   & 34 & 60 & 84\\
{\bf NLC-Value}   & 41 & 65 &86\\
\Xhline{1.5 pt}
\end{tabular}
\end{center}
\caption{the number of terms found in agrovoc foreach measure}
\label{tab3} 
\end{table}

\begin{table}[!h]
\begin{center}
\renewcommand {\arraystretch }{1}
\begin{tabular}{@{}lllD{,}{,}{1}D{,}{,}{1}@{}}
\Xhline{1.0 pt}
 & \multicolumn{4}{l}{{\bf Top MWT considred}}  \\  
\cline{2-5}
\centering {\bf Stat. measures} & \centering {\bf 100} & \centering {\bf 200}& \centering {\bf 300} & \tabularnewline
\hline
{\bf LLR} &  40 & 81 & 113\\
{\bf C-value} & 44 & 79 & 120\\
{\bf NC-value} & 42 & 78 & 123\\
{\bf NTC-value}& 45 & 83 & 126\\
{\bf LLR+C-value}   & 39 & 84 &121\\
{\bf NLC-Value}   & 41 & 86 & 133\\
\Xhline{1.5 pt}
\end{tabular}
\end{center}
\caption{the number of terms found in IATE foreach measure}
\label{tab4} 
\end{table}
\section{Conclusion}

\paragraph{} In this work, we have presented a hybrid method for Arabic MWT acquisition; this method takes advantage of existing linguistic and statistical approaches. As a first step, we apply linguistic filters to extract MWT candidates based on syntactic patterns using a sequence identifier component. Then, MWT variants are identified through a morphological analysis of he extracted MWTs based on light stemming. In the statistical step, we have proposed a novel statistical measure, $NLC$-value, that consists of ranking MWT candidates by considering contextual information and both termhood and unithood statistical measures.

\paragraph{}  Experiments are performed for bi-grams and tri-grams on an environment Arabic corpus. The experimental results show that our method outperforms the previous ones in term of quality of the extracted MWTs. In conclusion, the combination of the best association measures that integrate contextual information and both termhood and unithood statistical measures improves the performance of the MWT acquisition process.

\paragraph{} In a near future, we plan on using the extracted MWTs in an information retrieval system as complex terms often constitute a better representation of the content of a document than single word terms.

\end{document}